\documentclass[10pt,twocolumn,letterpaper]{article}

\usepackage[pagenumbers]{cvpr}
\usepackage{multirow}

\definecolor{posgreen}{RGB}{0,140,60}
\definecolor{negred}{RGB}{200,30,30}
\newcommand{\gain}[1]{\textcolor{posgreen}{#1}}
\newcommand{\drop}[1]{\textcolor{negred}{#1}}

\usepackage{xspace}
\newcommand{\dataset}{SORBET\xspace}

\newcommand{\method}{Nunchi\xspace}

\graphicspath{{figures/}}

\usepackage{booktabs}
\usepackage{multirow}
\usepackage{placeins}
\usepackage{makecell}
\usepackage{pdflscape}
\usepackage{microtype}

\usepackage{adjustbox}
\usepackage{amssymb}
\usepackage{pifont}
\newcommand{\cmark}{\ding{51}}
\newcommand{\xmark}{\ding{55}}

\renewcommand{\paragraph}[1]{\vspace{.5em}\noindent\textbf{#1.}}

\usepackage{xspace}

\definecolor{cvprblue}{rgb}{0.21,0.49,0.74}
\usepackage[pagebackref,breaklinks,colorlinks,allcolors=cvprblue]{hyperref}

\title{Nunchi: Ears Help Eyes in Multi-Party Conversations}

\author{
JunYeong Park$^{1}$ \quad Shijing Wang$^{2}$ \quad Seonwook Park$^{3}$ \quad Michael Stengel$^{3}$\\
Shalini De Mello$^{3}$ \quad Sang-Il Choi$^{1}$\thanks{Corresponding author.} \quad Hyung Jin Chang$^{4}$\\[4pt]
$^1$Dankook University \quad $^2$Beijing Jiaotong University \quad $^3$NVIDIA \quad $^4$University of Birmingham
}

\begin{document}
\maketitle
\begin{abstract}
Identifying whom a person looks at is central to understanding multi-party conversations, yet gaze geometry alone becomes ambiguous as more people appear.
Conversations offer cues that geometry lacks, such as who is speaking and what is said, but no gaze dataset provides them for many multi-party scenes.
We introduce \textbf{SORBET}, a benchmark of 3,340 samples from 1,578 clips with synchronized speech, transcripts, and attended person labels, covering more multi-party scenes than existing gaze datasets.
Prior gaze models struggle to use these cues, which come in different forms, must be linked to the right person, and can conflict with gaze geometry.
We propose \textbf{Nunchi}, a framework that constructs a text prompt from the transcript and the outputs of gaze and speaker models, links every cue to the candidates with shared labels, and lets a vision-language model (VLM) weigh gaze geometry against conversational cues to identify the attended person.
\method outperforms the strongest existing method by 4.8 percentage points.
On SORBET, conversational cues help whereas additional visual cues do not, and the gains hold across three VLMs, with automatic transcripts, and on two existing datasets.
\end{abstract}

\section{Introduction}
\label{sec:intro}

Whom each person looks at is central to understanding conversations among several people~\cite{KENDON196722, Otsuka2005}.
Recognizing this social attention benefits applications such as education, healthcare, and human-robot interaction.
Existing gaze tasks address this question only indirectly.
Gaze following predicts where a person looks~\cite{recasensWhereAreThey2015}, and social gaze predicts relations such as looking at humans (LAH)~\cite{guptaMTGSNovelFramework2024}.
We formulate \emph{attended person identification}: given an \emph{observer} and the other people in the scene as \emph{candidates}, identify whom among the candidates the observer looks at, \ie, the \emph{attended person}.
We focus on multi-party conversations of three or more people~\cite{gu2022says}, where gaze alone is often ambiguous but the conversation offers cues.

\begin{figure}[t]
  \centering
  \includegraphics[width=\linewidth]{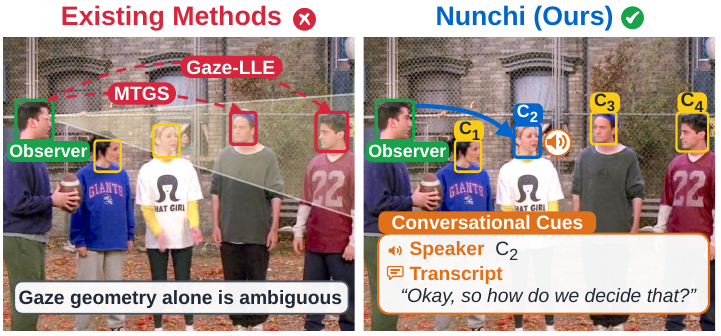}
  \caption{\textbf{Attended person identification in a multi-party conversation.}
  As more people appear in a scene, several candidates lie in a similar direction from the observer, so gaze geometry alone cannot determine whom the observer looks at.
  Gaze-LLE~\cite{ryanGazeLLEGazeTarget2025} and MTGS~\cite{guptaMTGSNovelFramework2024} each select an incorrect candidate.
  \method combines gaze geometry with conversational cues, the speaker and the transcript, and identifies the attended person $C_2$.}
  \label{fig:teaser}
  \vspace{-3mm}
\end{figure}

Existing methods can identify the attended person in two ways.
Gaze following methods such as Gaze-LLE~\cite{ryanGazeLLEGazeTarget2025} predict the gaze target as a location on a heatmap~\cite{recasensWhereAreThey2015, tafascaSharinganTransformerArchitecture2024, tafascaChildPlayNewBenchmark2023}, from which the attended person can be selected by the heatmap mass or peak.
Social gaze methods such as MTGS~\cite{guptaMTGSNovelFramework2024} predict the LAH probability of each person.
Both rely on gaze geometry, which relates the gaze direction of the observer to the positions of the candidates.
They perform well when few people are present.
However, their accuracy decreases as more people appear in the scene, as shown in \cref{tab:k}.
On VAT~\cite{chong2020detecting}, for example, the accuracy of Gaze-LLE drops from 83.9\% with two other people to 64.3\% with five or more.
With more people, several of them often lie in a similar direction from the observer.
A small error in the estimated direction then changes the selected person.
\cref{fig:teaser} shows such a case.
Gaze geometry alone is thus often ambiguous when many people are present.

Conversations provide cues that gaze geometry lacks~\cite{Otsuka2005}.
We consider two conversational cues~\cite{gu2022says}, the speaker cue for who is speaking and the transcript cue for what is said.
Listeners often look at the current speaker~\cite{KENDON196722}.
A speaker may also name the observer, as in ``Ross, come here,'' which draws the attention of the observer to the speaker.
Prior gaze models have used the speaker~\cite{guptaMTGSNovelFramework2024, houMultiModalGazeFollowing2024}, but with limited success.
We attribute this to two factors, the lack of multi-party gaze datasets with conversational cues and the way the cues are used.
As shown in \cref{tab:datasets}, no existing gaze dataset combines video, synchronized speech, and transcripts with many multi-party samples.

We therefore build \dataset (SOcial gaze Recordings with Both spEech and Transcripts), a dataset of 3,340 samples of multi-party conversations from three TV series, each with synchronized speech, a transcript, and an attended person label.
Its keyframes come from 1,578 clips across 241 episodes, and each shows at least three people and the speaker, so the samples cover diverse multi-party scenes.
\dataset thus allows us to train models with conversational cues and to measure how much each cue helps.

Even with such data, combining these cues with gaze raises three core challenges related to multimodal learning~\cite{liang2024foundations}.
First, the cues are heterogeneous~\cite{liang2024foundations}.
Gaze geometry is a heatmap over the image, the speaker cue is a score per person, and the transcript is a sequence of utterances.
Fusing them at the feature level brings little gain, as MTGS~\cite{guptaMTGSNovelFramework2024} reports for its speaking-status input.
Second, the cues are not automatically aligned with the people in the image.
A voice or a name helps only when it is linked to the right face.
Third, the cues can conflict.
Gaze geometry may favor one candidate while the speaker cue favors another, so the model must reason over both.

\begin{table}[t]
\centering
\small
\setlength{\tabcolsep}{4pt}
\begin{tabular}{@{}lcccccc@{}}
\toprule
 & \multicolumn{2}{c}{$K{=}2$} & \multicolumn{2}{c}{$K{=}3$--$4$} & \multicolumn{2}{c}{$K{\geq}5$} \\
\cmidrule(lr){2-3}\cmidrule(lr){4-5}\cmidrule(l){6-7}
Dataset & G & M & G & M & G & M \\
\midrule
VAT$^\dagger$         & 83.9 & 85.1 & 80.2 & 77.0 & 64.3 & 63.8 \\
ChildPlay$^\dagger$   & 87.7 & 86.9 & 81.3 & 79.2 & 57.1 & 67.3 \\
UCO-LAEO$^\dagger$    & 100.0 & 97.4 & 96.2 & 100.0 & -- & -- \\
VGS                   & 79.3 & 75.9 & 55.9 & 58.8 & -- & -- \\
\bottomrule
\end{tabular}
\caption{\textbf{Effect of the number of people on Gaze-LLE (G) and MTGS (M)}, in identity accuracy (\%).
Both models use their released checkpoints.
$K$ is the number of annotated people other than the observer.
$^\dagger$ marks datasets used to train MTGS.
}
\label{tab:k}
\vspace{-3mm}
\end{table}

To this end, we propose \method\footnote{\textit{Nunchi} is a Korean word for reading the room, where \textit{nun} means eye. A person with \textit{nunchi} knows who is paying attention to whom by watching where people look and listening to who is talking and what they say. Our model does the same.}, a framework that addresses each challenge with a dedicated design.
For heterogeneity, \method writes expert outputs as text in a shared language space.
VLMs estimate gaze direction less accurately than dedicated gaze models~\cite{wang2025vl4gaze, wang2026eyes}, likely because they are trained for semantic understanding rather than precise spatial geometry.
We therefore use a frozen Gaze-LLE to estimate the gaze geometry, and call its output the gaze prior.
A frozen active speaker detector provides the speaker cue, and the transcript what is said.
For alignment, we draw each candidate's face box with a label $C_k$ on the image as a visual prompt~\cite{yang2023setofmark}, and use the same labels in every cue.
For reasoning under conflict, a VLM reads the gaze prior and the conversational cues together and scores every candidate in a single forward pass.
Unlike fusion modules that mix features, this design lets the model compare gaze geometry and conversational cues explicitly for each sample.

\method reaches 86.7\% on \dataset, 4.8 points above the strongest existing method.
With five or more candidates, it is 8.0 points above the best baseline.
Given the same speaker scores, \method gains 1.9 points, whereas MTGS gains 0.8.
The conversational cues complement the gaze prior across three VLMs.
In contrast, additional visual cues, such as depth and mutual gaze, do not improve accuracy, whether given as text or drawn on the image.
These gains hold with automatic transcripts and on two existing datasets.

Our contributions are as follows.
\begin{enumerate}
  \item We formulate attended person identification in multi-party conversations as a choice among the candidates, and show that gaze alone becomes ambiguous as more people appear, while conversational cues reduce this ambiguity.
  \item We build SORBET, to our knowledge the first benchmark for identifying the attended person in multi-party conversations with synchronized speech and transcripts. It contains 3,340 samples from 1,578 clips across diverse scenes, and provides attended person labels, an evaluation protocol, and results of gaze following, social gaze, and VLM baselines. We will release the annotations.
  \item We propose \method, a framework that identifies the attended person by combining gaze with speech and transcripts. It writes the expert outputs as text, ties every cue to the candidates with shared labels, and scores the candidates with a single VLM.
\end{enumerate}

\begin{table}[t]
\centering
\small
\setlength{\tabcolsep}{4pt}
\resizebox{\columnwidth}{!}{
\begin{tabular}{@{}lcccrr@{}}
\toprule
 & & & & \multicolumn{2}{c}{$K{\ge}3$ (test)} \\
\cmidrule(l){5-6}
Dataset & Video & Audio & Transcript & Clips & Samples \\
\midrule
GazeFollow~\cite{recasensWhereAreThey2015} & \xmark & \xmark & \xmark & -- & 580 \\
VAT~\cite{chong2020detecting}                  & \cmark & \xmark & \xmark & 32 & 383 \\
ChildPlay~\cite{tafascaChildPlayNewBenchmark2023}    & \cmark & \xmark & \xmark & 9 & 97 \\
UCO-LAEO~\cite{marin2019laeo_uco}         & \cmark & \cmark & \xmark & 5 & 26 \\
AVA-LAEO~\cite{marin2019laeo_uco}         & \cmark & \xmark & \xmark & 10$^{*}$ & 110$^{*}$ \\
VGS~\cite{houMultiModalGazeFollowing2024}                 & \cmark & \cmark & \xmark & 1 & 34 \\
\midrule
SORBET (ours)                            & \cmark & \cmark & \cmark & 169 & 508 \\
\bottomrule
\end{tabular}%
}
\caption{\textbf{Comparison with existing gaze following and social gaze datasets.}
Audio and Transcript mark synchronized audio and transcripts released with the dataset.
$K{\geq}3$ counts test clips and samples with three or more candidates.
As adjacent video frames are nearly identical, we sample video datasets every 25 frames, about once per second, and SORBET at keyframes.
$^*$No test split, so we count the validation movies.}
\label{tab:datasets}
\vspace{-3mm}
\end{table}
\section{Related Work}
\label{sec:related}

\paragraph{Gaze Following}
Gaze following predicts where a person looks in an image~\cite{recasensWhereAreThey2015,chong2018connecting}, with extensions to video~\cite{chong2020detecting}.
Several methods incorporate depth or 3D cues~\cite{fang2021dual_depth, Bao2022escnet_depth, Horanyi2023where_depth}, or object-level information~\cite{tonini2023object,miMultiscaleObjectAwareGaze2026} to separate targets at different distances.
Sharingan~\cite{tafascaSharinganTransformerArchitecture2024} efficiently predicts multiple people's targets at once with per-person gaze tokens.
Gaze-LLE~\cite{ryanGazeLLEGazeTarget2025} shows that a frozen general-purpose encoder with a lightweight decoder achieves state-of-the-art results; we adopt it as our gaze expert.
These methods output a location, and selecting the attended person from it becomes ambiguous when several people lie in a similar direction.
In contrast, \method resolves such cases with conversational cues.

\paragraph{Social Gaze and Visual Focus of Attention}
Social gaze methods detect mutual gaze in videos~\cite{marin2019laeo_uco} and images~\cite{doosti2021boosting_mutualgaze}, gaze communication~\cite{fan2019understanding_graph}, and joint attention~\cite{nakatani2023interaction}.
Gupta~\etal~\cite{gupta2024unified} and MTGS~\cite{guptaMTGSNovelFramework2024} jointly predict gaze targets and social gaze relations, including LAH.
MTGS processes multiple people over time, and its VSGaze dataset unifies several video datasets with social gaze labels.
Meeting studies estimate the visual focus of attention (VFOA) of each participant among a fixed set of targets from head pose~\cite{baRecognizingVisualFocus2009,Stiefelhagen2002}.
Ba and Odobez~\cite{ba2011multiperson} add speaking status as context, and Otsuka \etal~\cite{Otsuka2005} use utterances to infer conversation structure.
These studies consider about four participants at fixed seats.
Attended person identification is a person-level form of VFOA, which we study from a single keyframe of TV conversations with three to more than six people in varying arrangements.

\paragraph{Gaze Datasets}
GazeFollow~\cite{recasensWhereAreThey2015} provides gaze targets in images, and VAT~\cite{chong2020detecting} and ChildPlay~\cite{tafascaChildPlayNewBenchmark2023} extend them to videos, including targets outside the frame and children.
UCO-LAEO and AVA-LAEO~\cite{marin2019laeo_uco} label people looking at each other in TV shows and movies, and VGS~\cite{houMultiModalGazeFollowing2024} brings audio to gaze following in conversations.
As shown in \cref{tab:datasets}, none of them provides transcripts, and few of their test samples have three or more candidates.
Multi-party conversation datasets~\cite{wang2025friends, zhang2024mintrec2} provide speech, transcripts, and speaker labels but no gaze labels, which SORBET adds.

\paragraph{Conversational Cues and VLMs for Gaze Understanding}
MMGaze~\cite{houMultiModalGazeFollowing2024} introduces audio into gaze following in conversations, on videos with up to four people.
MTGS~\cite{guptaMTGSNovelFramework2024} accepts a speaking-status input and reports little LAH change from it.
Active speaker detection finds who is speaking from audio and face tracks~\cite{roth2020ava_asd,Liao_lrasd_2023_CVPR}, but not whom the others look at, and we use it as a cue extractor.
Lee~\etal~\cite{lee2024modeling} align utterances with the visual features of each player to identify whom a speaker addresses in social deduction games, and Ouyang~\etal~\cite{ouyang2026multi} align them inside the attention of a multimodal LLM, but neither considers gaze targets.
VL4Gaze~\cite{wang2025vl4gaze} and Eyes on VLM~\cite{wang2026eyes} evaluate how well VLMs understand gaze, including gaze following and social gaze.
GazeLLM~\cite{yang2025gazellm} reasons over structured scene representations to find gaze targets, and GTA~\cite{nasiriboukani2026gaze_arrow} draws the predicted gaze as an arrow and asks a VLM to name the attended object.
These works use only the image and identify objects, while Lee and Deng~\cite{lee2025speakeraware} predict gaze directions in three-person conversations from transcripts without images.
To our knowledge, no prior gaze model combines transcripts with the image.
\method gives a VLM the gaze prior and conversational cues as text and identifies the attended person.

\section{The SORBET Dataset}
\label{sec:data}
\begin{figure*}[t]
  \centering
  \includegraphics[width=\textwidth]{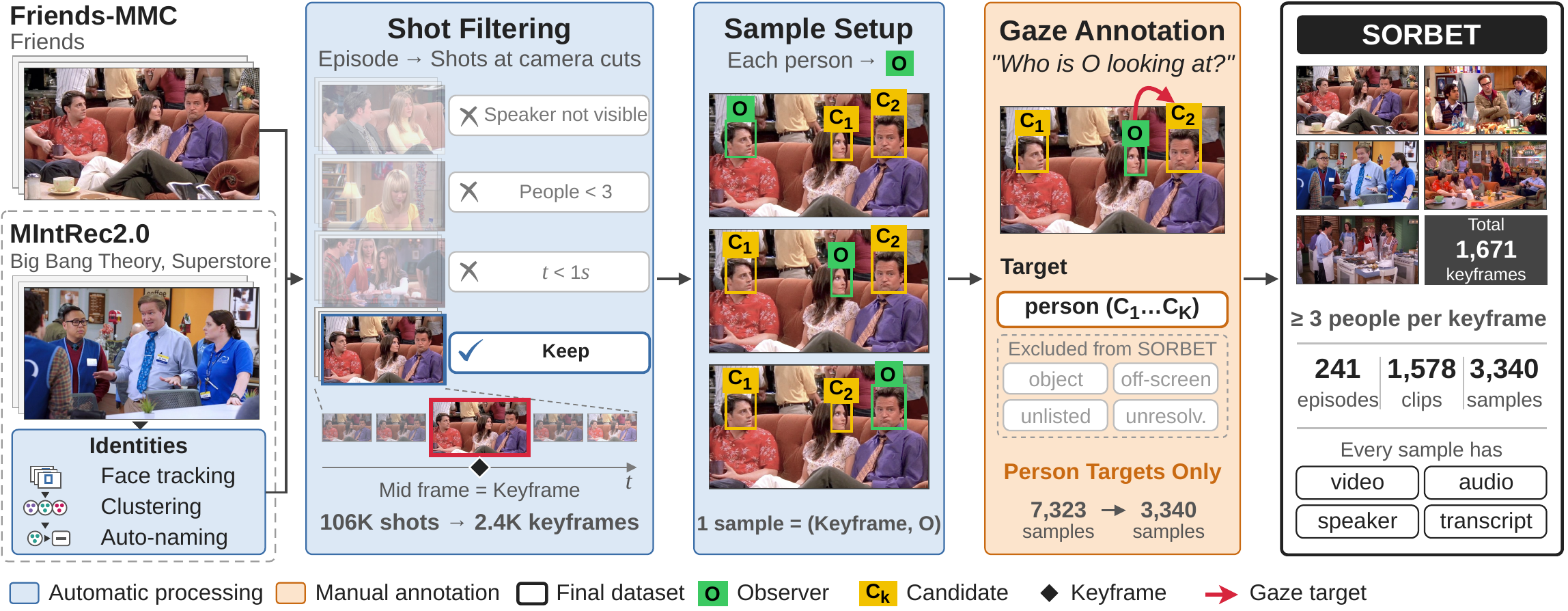}
  \caption{\textbf{Construction of SORBET.}
  From two multi-party conversation datasets, we keep frames with at least three identified people and a visible speaker, and take the middle frame of each segment of such frames as a keyframe.
  Each identified person serves in turn as the observer, and the annotator labels what the observer looks at.
  SORBET keeps samples whose gaze target is a candidate.}
  \label{fig:preprocess}
  \vspace{-3mm}
\end{figure*}

\subsection{Data Collection}
As shown in \cref{tab:datasets}, no existing gaze dataset provides synchronized speech and transcripts for conversations among three or more people.
We therefore extend two existing multimodal multi-party conversation datasets~\cite{wang2025friends, zhang2024mintrec2} with manually labeled attended persons.

\paragraph{Source Datasets}
Friends-MMC~\cite{wang2025friends} covers 220 episodes from 10 seasons of the TV series \textit{Friends}, with video clips, transcripts, face bounding boxes, character identities of faces, and speaker labels.
MIntRec2.0~\cite{zhang2024mintrec2} is a multimodal intent recognition dataset collected from \textit{Friends}, \textit{The Big Bang Theory} (BBT), and \textit{Superstore}, with video clips, transcripts, and speaker labels but no face bounding boxes or character identities.
As its \textit{Friends} subset overlaps with Friends-MMC, we use only BBT and \textit{Superstore}.
Both datasets provide a video clip with synchronized speech for each utterance, together with its transcript and speaker label, so the speech is diarized at the utterance level.
We call people whose faces have a character identity identified people.
Friends-MMC provides these identities, and we assign them for MIntRec2.0 as described below.

\paragraph{Preprocessing}
As shown in \cref{fig:preprocess}, we process both datasets in the same way.
MIntRec2.0 does not provide the episode of each clip, which we need to split the data at the episode level.
We therefore infer pseudo-episodes from its dialogue IDs, which group the utterances of one conversation, and from utterance timestamps, as detailed in the supplementary material.
PySceneDetect~\cite{Castellano_PySceneDetect} detects shot boundaries in both datasets.

\paragraph{Identities in MIntRec2.0}
We build face tracks within each shot and cluster them by face embeddings.
We assign a character name to each cluster from the speaker labels of clips that contain a single visible face.
The resulting identities agree with LR-ASD~\cite{Liao_lrasd_2023_CVPR}, an independent active speaker detection model, on 93--94\% of multi-face utterances whose speaker is among the identified people.
Faces in unnamed clusters are used neither as observers nor as candidates, and details are in the supplementary material.

\paragraph{Keyframe Selection}
We keep frames in which (a) at least three identified people are visible, so that every observer has at least two candidates, and (b) the speaker of the clip is visible, so that the speaker cue is defined for every sample.
Consecutive kept frames within a shot form a segment, and we take the middle frame of each segment of at least 24 frames, about one second, as a keyframe.
One keyframe per segment avoids near-duplicate samples.
Each identified person in a keyframe serves in turn as the observer, and the other identified people are their candidates.
A sample is thus a (keyframe, observer) pair with $K$ candidates, and we obtain 7,323 samples from 2,410 keyframes for annotation.

\subsection{Data Annotation}
\paragraph{Annotation Protocol}
Our Gradio-based interface~\cite{abid2019gradio} shows the keyframe with face boxes, the surrounding clip with audio, and the transcript of the utterance, but no model predictions such as gaze heatmaps or speaker scores.
Each sample is labeled as stable with one target, as shift when the gaze moves between two targets at the keyframe, or as unresolvable.
The target is a candidate, an object, something off-screen, or a person without a face box.
In the first pass, 45.8\% of samples have a candidate as the final target, as detailed in the supplementary material.

\paragraph{Label Consistency}
We measure agreement with Cohen's $\kappa$ within the annotator and Fleiss' $\kappa$ across annotators, both of which correct for chance agreement and equal 1 for perfect agreement.
The annotator re-labels 100 samples per TV series without seeing the earlier labels, giving $\kappa$ of 0.82, 0.93, and 0.87 for \textit{Friends}, BBT, and \textit{Superstore}.
The annotator and five others also label 100 common test samples.
Another annotator who chooses a candidate picks the annotator's candidate in 93.6\% of cases.
On the 57 samples where all six choose a candidate, Fleiss' $\kappa$ is 0.95.
Most disagreements concern whether the target is a candidate at all, as 77 of the 104 disagreeing labels mark an object, something off-screen, a gaze shift, or an unresolvable target.
\Cref{sec:existing} also tests the cues on datasets labeled by others, and agreement over all label types is in the supplementary material.

\paragraph{Sample Selection}
\label{sec:subset}
SORBET keeps only samples whose final target is a candidate.
Whether the observer looks at a person is the LAH task~\cite{guptaMTGSNovelFramework2024}, and we address the next question, whom among the candidates the observer looks at.
Rules for shift samples are in the supplementary material.
The resulting SORBET dataset contains 3,340 samples from 241 episodes.
As shown in \cref{tab:split}, we split it at the episode level into 1,839 training, 658 validation, and 843 test samples.
The attended person is the current speaker in 57.8\% of samples, so the speaker alone does not determine the answer.

\begin{table}[t]
  \centering
  \footnotesize
  \setlength{\tabcolsep}{2pt}
  \begin{tabular*}{\columnwidth}{@{\extracolsep{\fill}}llrrrrrrr@{}}
    \toprule
    Series & Split & Episode & Sample & $K{=}2$ & $K{=}3$ & $K{=}4$ & $K{\geq}5$ & Spk.\ (\%) \\
    \midrule
    \multirow{3}{*}{\textit{Friends}}    & train & 108 & 1,118 & 328 & 540 & 197 & 53 & 49.6 \\
                                         & val   &  42 &   485 & 143 & 278 &  60 &  4 & 48.5 \\
                                         & test  &  27 &   480 & 108 & 220 &  74 & 78 & 50.8 \\
    \midrule
    \multirow{3}{*}{BBT}                 & train &  16 &   219 & 134 &  65 &  20 &  0 & 64.8 \\
                                         & val   &   2 &    48 &  35 &  13 &   0 &  0 & 58.3 \\
                                         & test  &   5 &   139 &  94 &  40 &   5 &  0 & 72.7 \\
    \midrule
    \multirow{3}{*}{\textit{Superstore}} & train &  29 &   502 & 325 & 102 &  38 & 37 & 73.5 \\
                                         & val   &   4 &   125 &  66 &  23 &  19 & 17 & 72.0 \\
                                         & test  &   8 &   224 & 133 &  42 &  27 & 22 & 75.0 \\
    \midrule
    \multicolumn{2}{@{}l}{Total}         &       241 & 3,340 & 1,366 & 1,323 & 440 & 211 & 57.8 \\
    \bottomrule
  \end{tabular*}
  \caption{\textbf{SORBET splits}. Splits are disjoint at the episode level. $K$ is the number of candidates per sample. Spk.\ is the percentage of samples whose target is the current speaker.}
  \label{tab:split}
  \vspace{-3mm}
\end{table}

\section{Method}
\label{sec:method}
\begin{figure*}[t]
  \centering
  \includegraphics[width=\textwidth]{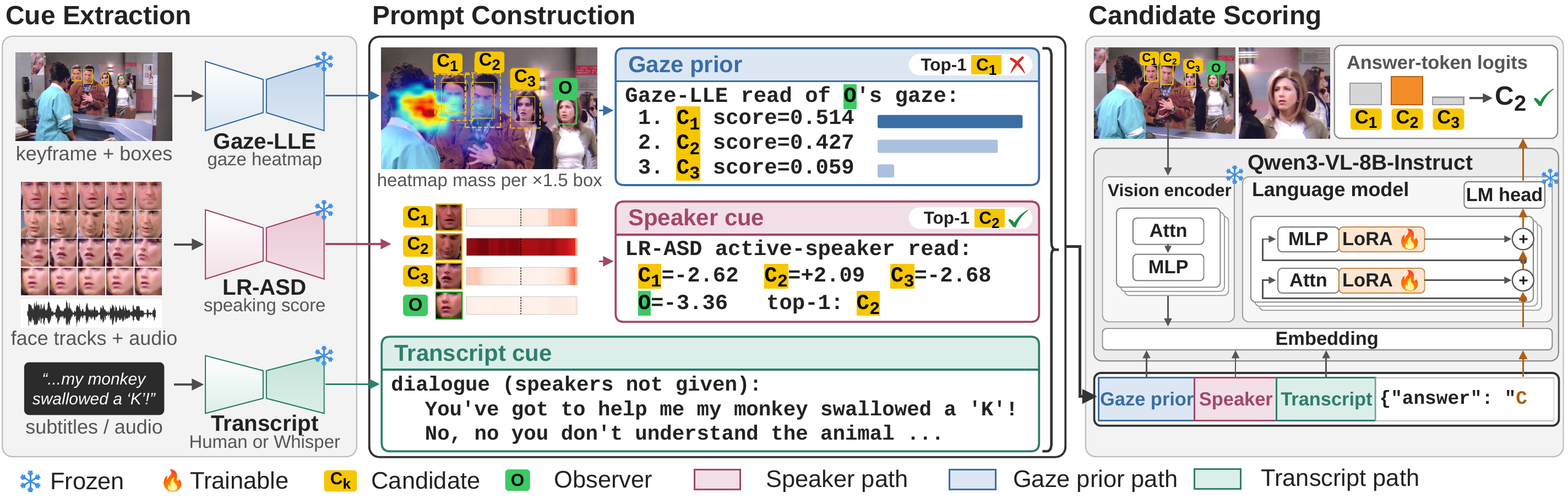}
  \caption{\textbf{Overview of \method.}
  Cue extraction obtains gaze and speaker information with frozen experts, along with the transcript.
  Prompt construction writes them as text, and every cue refers to each candidate by the label $C_k$ drawn in the keyframe.
  Candidate scoring reads the two images and the prompt with a single VLM and scores each candidate by the logit of its label.
  In this example, the gaze prior and the speaker cue point to different candidates, and \method selects the attended person $C_2$.
  Colors mark the gaze prior, speaker, and transcript paths, and \cmark/\xmark\ mark agreement with the ground truth, which the model does not see.}
  \label{fig:method}
  \vspace{-3mm}
\end{figure*}

\subsection{Overview}
\label{sec:overview}
Given a keyframe $I$, the face bounding box $O$ of an observer, and the face bounding boxes $\{C_k\}_{k=1}^{K}$ of $K$ candidates, we predict the index $\hat{y}\in\{1,\dots,K\}$ of the attended person.
As shown in \cref{fig:method}, \method works in three stages.
Cue extraction obtains gaze and speaker information with frozen expert models and collects the transcript.
Prompt construction writes the expert outputs as text and labels each candidate as $C_k$ in the image and in every cue.
Candidate scoring reads the images and the prompt with a single VLM.
Because all information enters through the input, cues can be added or removed without changing the architecture.

\subsection{Cue Extraction}
\label{sec:cue_extraction}
Gaze-LLE~\cite{ryanGazeLLEGazeTarget2025} (ViT-L, trained on GazeFollow~\cite{recasensWhereAreThey2015}) takes the keyframe and the face box of $O$ and outputs a gaze heatmap.
For each candidate, $m_k$ is the fraction of the heatmap mass inside its face box enlarged by $1.5\times$.
LR-ASD~\cite{Liao_lrasd_2023_CVPR}, an active speaker detection model trained on AVA-ActiveSpeaker~\cite{roth2020ava_asd}, predicts from the face tracks of \cref{sec:data} and the audio whether each person is speaking in each frame.
Averaging these scores over the current utterance gives $a_k$ for each candidate and $a_O$ for the observer, so a high $a_k$ means that $C_k$ is likely speaking.
We use the human-written transcripts of the source datasets by default, and Whisper~\cite{radford2023whisper} transcripts in \cref{sec:existing}.

\subsection{Prompt Construction}
\label{sec:prompt}
The VLM receives two images and a text prompt.
The first image is the keyframe with face boxes, which serves as a visual prompt~\cite{yang2023setofmark}.
$O$ is drawn in green, and each candidate is drawn in yellow with its label $C_k$.
The second image is a $448\times448$ observer crop.
The prompt states that $O$ looks at exactly one of the candidates and lists their labels.
The gaze prior and each cue then follow as separate paragraphs that refer to the candidates by the same labels.

\paragraph{Gaze Prior}
The candidates are listed in descending order of $m_k$, with whether the heatmap peak falls in each box and the distance from the peak to the box center.
The paragraph also gives the coordinates of the peak and of the face center of $O$, and the fraction of the mass outside all candidate boxes, which is large when the heatmap misses the candidates.
Its top-ranked candidate is the prediction of zero-shot Gaze-LLE.
The two images and this paragraph form the base input.

\paragraph{Speaker Cue}
The speaker cue lists $a_O$ and all $a_k$, together with the top-scoring face in the frame.

\paragraph{Transcript Cue}
The transcript cue gives the current utterance and the two preceding ones, each with its start offset in frames from the keyframe.
We remove the speaker name of each utterance, so that speaker information comes only from the speaker cue.
We also replace the names of $O$ and the candidates with their labels, using the character identities of \cref{sec:data}, \eg, ``Ross, come here'' becomes ``$\text{C}_2$, come here''.
The supplementary material shows a complete prompt.

\subsection{Candidate Scoring and Training}
\label{sec:scoring}
We use Qwen3-VL-8B-Instruct~\cite{baiQwen3VLTechnicalReport2025} as the VLM and append the prefix \verb|{"answer": "C| to the prompt, so that the next token is a candidate index.
The logit $z_k$ of index $k$ is the score of $C_k$, and $\hat{y}=\arg\max_k z_k$.
All scores come from one forward pass over the images and the prompt, so the VLM weighs the cues against each other when they disagree.
We fine-tune the language model with LoRA~\cite{huLoRALowRankAdaptation2021} and freeze the vision encoder.
The target is a JSON object with the answer first, followed by a short rationale that templates fill from the annotations, including the current speaker, as detailed in the supplementary material.
The loss covers all target tokens.
At inference, $z_k$ is read before any rationale token is generated, so the prediction does not depend on the generated rationale.

\section{Experiments}
\label{sec:exp}
\begin{table*}[t]
  \centering
  \footnotesize
  \begin{tabular*}{\textwidth}{@{\extracolsep{\fill}}llllllrrrrr@{}}
    \toprule
    & Model & Training & Output & Cue & Params (trainable) & All & \textit{Friends} & BBT & \textit{Superstore} & $K{\geq}5$ \\
    \midrule
    \multirow{6}{*}{\makecell[l]{Gaze\\following}}
      & \multirow{4}{*}{Gaze-LLE~\cite{ryanGazeLLEGazeTarget2025}}
        & zero-shot   & Mass & -- & 307.3M (0)      & 79.6 & 77.5 & 86.3 & 79.9 & 64.0 \\
      & & decoder     & Mass & -- & 307.3M (2.9M)   & 81.9 & 81.7 & 86.3 & 79.5 & 66.0 \\
      & & LoRA + dec. & Mass & -- & 309.6M (5.3M)   & 81.6 & 80.2 & 85.6 & 82.1 & 67.0 \\
      & & full        & Mass & -- & 307.3M (307.3M) & 79.8 & 78.1 & 87.1 & 79.0 & 64.0 \\
    \cmidrule(l){2-11}
      & \multirow{2}{*}{Sharingan~\cite{tafascaSharinganTransformerArchitecture2024}}
        & zero-shot & Mass & -- & 106.1M (0)    & 76.3 & 75.2 & 80.6 & 75.9 & 58.0 \\
      & & decoder   & Mass & -- & 106.1M (7.9M) & 76.4 & 75.0 & 80.6 & 76.8 & 57.0 \\
    \midrule
    Multimodal
      & MMGaze~\cite{houMultiModalGazeFollowing2024}
        & full & Box & S & 62.6M (62.4M) & 57.3 & 49.6 & 69.8 & 66.1 & 30.0 \\
    \midrule
    \multirow{3}{*}{\makecell[l]{Social\\gaze}}
      & \multirow{3}{*}{MTGS~\cite{guptaMTGSNovelFramework2024}}
        & zero-shot & LAH & -- & 155.4M (0)     & 75.2 & 74.4 & 81.3 & 73.2 & 57.0 \\
      & & full      & LAH & -- & 155.4M (67.6M) & 78.2 & 76.3 & 84.2 & 78.6 & 61.0 \\
      & & full      & LAH & S  & 155.4M (67.6M) & 79.0 & 77.1 & 84.9 & 79.5 & 61.0 \\
    \midrule
    \multirow{6}{*}{\makecell[l]{VLM\\(w/o\\Gaze-LLE\\evidence)}}
      & \multirow{2}{*}{Qwen3-VL-8B~\cite{baiQwen3VLTechnicalReport2025}}
        & LoRA & Logit & --  & 8.85B (87.3M) & 76.2 & 71.0 & 84.2 & 82.1 & 60.0 \\
      & & LoRA & Logit & S+T & 8.85B (87.3M) & 81.7 & 75.6 & 90.7 & 89.3 & 68.0 \\
    \cmidrule(l){2-11}
      & \multirow{2}{*}{InternVL3.5-8B~\cite{wang2025internvl3.5}}
        & LoRA & Logit & --  & 8.62B (87.3M) & 72.7 & 69.8 & 82.7 & 72.8 & 54.0 \\
      & & LoRA & Logit & S+T & 8.62B (87.3M) & 79.4 & 73.3 & 88.5 & 86.6 & 67.0 \\
    \cmidrule(l){2-11}
      & \multirow{2}{*}{Qwen2.5-Omni-7B~\cite{xuQwen25OmniTechnicalReport2025}}
        & LoRA & Logit & --  & 9.01B (80.7M) & 70.3 & 66.0 & 77.7 & 75.0 & 55.0 \\
      & & LoRA & Logit & S+T & 9.01B (80.7M) & 75.3 & 68.5 & 83.5 & 84.8 & 60.0 \\
    \midrule
    \multirow{3}{*}{\method (Ours)}
      & \multirow{3}{*}{\makecell[l]{Qwen3-VL-8B\\+ gaze prior}}
        & LoRA & Logit & --  & 8.85B (87.3M) & 83.3 & 79.8 & 88.5 & 87.5 & 67.0 \\
      & & LoRA & Logit & S   & 8.85B (87.3M) & 85.2 & 81.9 & 89.9 & 89.3 & 74.0 \\
      & & LoRA & Logit & S+T & 8.85B (87.3M) & \textbf{86.7} & \textbf{82.9} & \textbf{92.1} & \textbf{91.5} & \textbf{76.0} \\
    \bottomrule
  \end{tabular*}
  \caption{\textbf{Comparison with state-of-the-art methods on the SORBET test set}, in identity accuracy (\%).
  The Training column lists the parts fine-tuned on SORBET, all starting from the released checkpoints except MMGaze, which follows its original recipe.
  For each baseline, we report the best of several configurations.
  All configurations, including the frozen parts in each setting, are in the supplementary material.
  S and T denote the speaker and transcript cues, and Output follows the prediction rules in \cref{sec:setup}.
  Frozen Gaze-LLE and LR-ASD are not counted in our parameters.
  }
  \label{tab:sota}
  \vspace{-3mm}
\end{table*}

\subsection{Experimental Setup}
\label{sec:setup}
\paragraph{Datasets}
We train our model on SORBET and evaluate it on SORBET and, without further training, on two existing datasets, VGS~\cite{houMultiModalGazeFollowing2024} and AVA-LAEO~\cite{marin2019laeo_uco}.
We choose them because their official videos include audio, from which the speaker and transcript cues are obtained.
We do not use UCO-LAEO~\cite{marin2019laeo_uco}, whose videos partly come from the same TV series as SORBET.
VGS consists of conversational videos with gaze target annotations, and we use all 266 samples from its 11 videos.
AVA-LAEO provides looking-at-each-other labels on AVA movies~\cite{gu2018ava}, and the candidate face tracks come from AVA-ActiveSpeaker~\cite{roth2020ava_asd}.
As it has no official test split, we evaluate on 322 samples from the AVA validation movies.
For an in-domain comparison, we also train a separate model on 601 training and 174 validation samples from the AVA training movies.
Samples from both datasets follow SORBET, so the observer looks at one of two or more candidates, but the speaker need not be visible.
As neither dataset provides transcripts, we generate them with Whisper~\cite{radford2023whisper} and keep names as spoken.

\paragraph{Baselines}
We compare against existing models that can identify the attended person.
Gaze-LLE~\cite{ryanGazeLLEGazeTarget2025} and Sharingan~\cite{tafascaSharinganTransformerArchitecture2024} are heatmap-based gaze following models.
MTGS~\cite{guptaMTGSNovelFramework2024} is a social gaze model that predicts each person’s LAH probability, and we also provide its official speaking-status input with the same LR-ASD scores as ours.
MMGaze~\cite{houMultiModalGazeFollowing2024} is a multimodal gaze following model that takes the image and audio and estimates the speaker with its own module. %
Each model is evaluated with its released checkpoint and after fine-tuning on SORBET with its official recipe and our training and validation sets.
We also fine-tune Qwen3-VL-8B~\cite{baiQwen3VLTechnicalReport2025}, InternVL3.5-8B~\cite{wang2025internvl3.5}, and Qwen2.5-Omni-7B~\cite{xuQwen25OmniTechnicalReport2025} on our input without the gaze prior, that is, the two images with or without the conversational cues.\footnote{We use the official code and checkpoints released by the authors for all baselines, including the VLMs, except MMGaze, which we train on SORBET from ImageNet initialization because this gives higher accuracy than fine-tuning its released checkpoint. Its top-scoring box overlaps no candidate face in 26.2\% of the test samples. All MMGaze configurations are in the supplementary material.}

\paragraph{Evaluation Metric}
We report identity accuracy, the top-1 accuracy over the candidates.
We do not use the LAH F1 of MTGS~\cite{guptaMTGSNovelFramework2024}, since its precision is always 1 when every sample has a person target.
Gaze-LLE and Sharingan select the candidate with the largest heatmap mass in its $1.5\times$ enlarged face box, and selecting by the heatmap peak gives similar results in the supplementary material.
MTGS selects the candidate with the highest LAH probability, MMGaze the candidate matching its top-scoring detected box, and VLMs the candidate with the highest logit $z_k$ described in \cref{sec:scoring}.

\paragraph{Implementation Details}
Our model is built on Qwen3-VL-8B-Instruct and implemented in PyTorch~\cite{paszke2019pytorch} on a single NVIDIA A100 GPU.
We apply LoRA (rank 32, $\alpha{=}64$) to the language model and keep the vision encoder frozen.
We train with AdamW (learning rate $10^{-4}$) and an effective batch size of 8 for at most two epochs, selecting the epoch on 400 fixed validation samples.
All VLMs, including the baselines, use the same settings.

\subsection{Comparison with State-of-the-Art Methods}
\label{sec:sota}

\begin{figure*}[t]
  \centering
  \includegraphics[width=\textwidth]{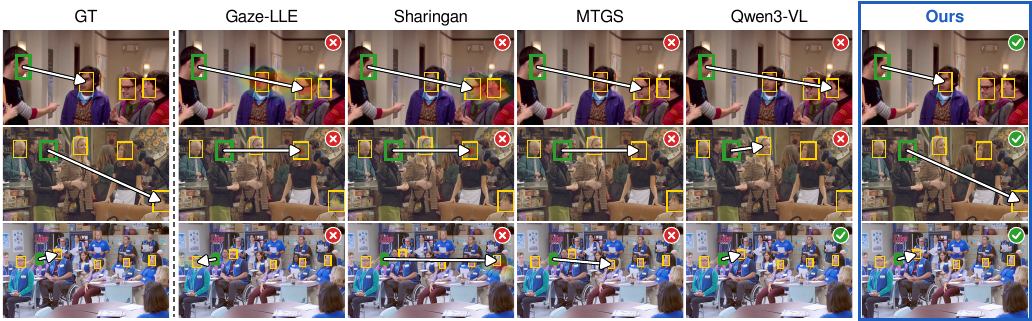}
  \caption{\textbf{Qualitative comparison on the SORBET test set.}
  Each column shows the candidate predicted by a method, with an arrow from the observer $O$ (green) to the prediction (yellow).
  Green and red marks indicate correct and incorrect predictions.
  We show examples where the gaze following and social gaze models fail and our model succeeds.}
  \label{fig:qual_sota}
  \vspace{-3mm}
\end{figure*}

\Cref{tab:sota} compares our model with gaze following, social gaze, and multimodal gaze following models, and with VLMs.
With the speaker and transcript cues, our model reaches 86.7\%, the highest accuracy among them overall and on each of the three series.
It outperforms the strongest existing method, Gaze-LLE with its decoder fine-tuned on SORBET, by 4.8 points.
The gain does not come from model size, since the same Qwen3-VL-8B without the gaze prior stays below the fine-tuned Gaze-LLE even with both cues.
With five or more candidates, it reaches 76.0\%, 8.0 points above the next best method, Qwen3-VL with the conversational cues but without the gaze prior.
The conversational cues add 9.0 points there, against 3.4 points over all samples.
The cues also raise the accuracy of all three VLMs without the gaze prior.
\Cref{fig:qual_sota} shows such cases.
The speaker and transcript cues thus help most when many people are present.

How the cues are given also matters.
MTGS receives the same LR-ASD scores as our speaker cue but gains 0.8 points from them, whereas our model gains 1.9 points.
MMGaze estimates the speaker with its own module and reaches 57.3\%.
Both fuse the speaker information with visual features inside their networks, whereas our model reads the speaker scores as text next to the gaze prior.
As a simpler way to combine the same scores, we replace the VLM with a logistic regression on $m_k$ and $a_k$ of each candidate.
As detailed in the supplementary material, it reaches 83.9\%, below the 85.2\% of our model with the speaker cue, and it cannot use the transcript.
When the gaze prior and the speaker cue point to different candidates, the gap over the logistic regression grows from 2.8 to 7.4 points, since \method weighs the cues per sample instead of with fixed weights.
The benefit of the speaker cue thus depends on how a model combines it with gaze geometry.

\begin{table}[!t]
  \centering
  \footnotesize
  \setlength{\tabcolsep}{3pt}
  \begin{tabular}{@{}lccccc@{}}
    \toprule
    Cue & All & \textit{Friends} & BBT & \textit{Superstore} & $K{\geq}5$ \\
    \midrule
    base input & 83.3 & 79.8 & 88.5 & 87.5 & 67.0 \\
    \midrule
    \multicolumn{6}{@{}l}{\textit{Conversational}} \\
    \hspace{0.6em}+ speaker              & 85.2 {\scriptsize (\gain{+1.9})}   & 81.9 & 89.9 & 89.3 & 74.0 \\
    \hspace{0.6em}+ transcript           & 83.2 {\scriptsize (\drop{$-$0.1})} & 79.0 & 89.9 & 88.0 & 64.0 \\
    \hspace{0.6em}+ speaker + transcript & \textbf{86.7} {\scriptsize \textbf{(\gain{+3.4})}} & \textbf{82.9} & \textbf{92.1} & \textbf{91.5} & \textbf{76.0} \\
    \midrule
    \multicolumn{6}{@{}l}{\textit{Visual}} \\
    \hspace{0.6em}+ reciprocal~\cite{doosti2021boosting_mutualgaze, guptaMTGSNovelFramework2024, marin2019laeo_uco} & 81.1 {\scriptsize (\drop{$-$2.2})} & 77.9 & 86.3 & 84.8 & 65.0 \\
    \hspace{0.6em}+ graph~\cite{fan2019understanding_graph}      & 81.7 {\scriptsize (\drop{$-$1.6})} & 78.3 & 86.3 & 86.2 & 65.0 \\
    \hspace{0.6em}+ depth~\cite{fang2021dual_depth,Bao2022escnet_depth, Horanyi2023where_depth}      & 82.2 {\scriptsize (\drop{$-$1.1})} & 80.2 & 86.3 & 83.9 & 69.0 \\
    \hspace{0.6em}+ arrow~\cite{nasiriboukani2025ggvl_arrow, nasiriboukani2026gaze_arrow}      & 82.6 {\scriptsize (\drop{$-$0.7})} & 80.6 & 87.8 & 83.5 & 72.0 \\
    \bottomrule
  \end{tabular}
  \caption{\textbf{Cue ablation on the SORBET test set}, in identity accuracy (\%).
  Each row adds cues to the base input, the two images and the gaze prior.
  Numbers in parentheses are changes from the base input in points.
  Speaker + transcript is our full model.}
  \label{tab:cue}
  \vspace{-3mm}
\end{table}

\subsection{Gaze Geometry and Conversational Cues}
\label{sec:gain}
\begin{table}[!t]
  \centering
  \footnotesize
  \begin{tabular*}{\columnwidth}{@{\extracolsep{\fill}}cc ccc@{}}
    \toprule
    Gaze-LLE & S+T & Qwen3-VL & InternVL3.5 & Qwen2.5-Omni \\
    \midrule
    \xmark & \xmark & 76.2 & 72.7 & 70.3 \\
    \xmark & \cmark & 81.7 \gain{(+5.5)} & 79.4 \gain{(+6.7)} & 75.3 \gain{(+5.0)} \\
    \cmark & \xmark & 83.3 \gain{(+7.1)} & 81.6 \gain{(+8.9)} & 80.6 \gain{(+10.3)} \\
    \cmark & \cmark & \textbf{86.7} \gain{(+10.5)} & \textbf{86.2} \gain{(+13.5)} & \textbf{84.6} \gain{(+14.3)} \\
    \bottomrule
  \end{tabular*}
  \caption{\textbf{Effect of the gaze prior and the conversational cues on the SORBET test set}, in identity accuracy (\%).
  The Gaze-LLE column marks the gaze prior in the input, and S+T marks the speaker and transcript cues.
  Numbers in parentheses are gains over the first row, which uses neither, in points.
  All models are fine-tuned with LoRA under the same setting.}
  \label{tab:gain}
\end{table}

\begin{figure*}[!t]
  \centering
  \includegraphics[width=\textwidth]{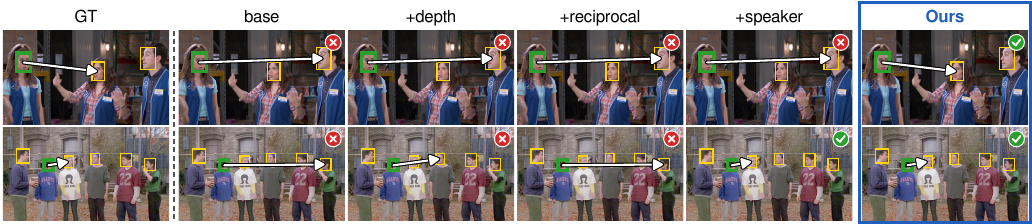}
  \caption{\textbf{Qualitative cue ablation on the SORBET test set.}
  All columns use Qwen3-VL-8B with the gaze prior (base), and each adds one cue.
  Green and red marks indicate correct and incorrect predictions.
  We show examples where the base input fails and our model succeeds.}
  \label{fig:qual_cue}
  \vspace{-3mm}
\end{figure*}

\Cref{tab:cue} tests whether the gain comes from conversational cues or simply from adding more cues.
We add each cue to the base input, the two images and the gaze prior of \cref{sec:prompt}.
Besides the speaker and transcript cues, we test four visual cues.
Depth gives the relative depth of $O$ and each candidate from Depth Anything V2~\cite{yang2024depth}.
Reciprocal gives the share of each candidate's Gaze-LLE heatmap that falls on the face of $O$.
Graph states whom each candidate looks at most, from the same per-candidate heatmaps.
Arrow adds a third image with an arrow from the face of $O$ to the peak of its heatmap.
Details and full results are in the supplementary material.

Combining the speaker and transcript cues gives the best accuracy, 3.4 points above the base input.
The speaker cue alone adds 1.9 points, while the transcript alone changes the accuracy little but adds 1.5 points on top of the speaker cue.
What is said thus becomes informative once the speaker is known.
In contrast, none of the visual cues improves the base input.
Three of them come from the same Gaze-LLE as the gaze prior and likely add little new information.
In \cref{fig:qual_cue}, depth and reciprocal keep the base input’s wrong prediction, whereas the conversational cues correct it.
Conversational cues thus provide information that gaze geometry cannot.

\begin{table}[!t]
  \centering
  \footnotesize
  \setlength{\tabcolsep}{4pt}
  \begin{tabular}{@{}llrrr@{}}
    \toprule
    Method & Train & SORBET & VGS & AVA-LAEO \\
    \midrule
    Chance                          & --         & 37.3 & 44.0 & 44.3 \\
    Gaze-LLE~\cite{ryanGazeLLEGazeTarget2025} (Mass)  & GazeFollow & 79.6 & 51.5 & 89.1 \\
    MTGS~\cite{guptaMTGSNovelFramework2024} (LAH)          & VSGaze     & 75.2 & 40.6 & 86.7 \\
    \midrule
    Ours                            & SORBET & 83.3 & 48.9 & 89.4 \\
    Ours + S                        & SORBET & 85.2 & 51.5 & 90.1 \\
    Ours + S + T$_\text{ASR}$       & SORBET & \textbf{86.4} \scriptsize{(86.7)} & \textbf{52.6} & \textbf{91.0} \\
    \midrule
    Ours                            & AVA-LAEO & -- & -- & 88.8 \\
    Ours + S + T$_\text{ASR}$       & AVA-LAEO & -- & -- & 89.4 \\
    \bottomrule
  \end{tabular}
  \caption{\textbf{Results on the SORBET, VGS, and AVA-LAEO test sets}, in identity accuracy (\%).
  The Train column gives the training data, and the baselines use their released checkpoints.
  S denotes the speaker cue, and T$_\text{ASR}$ denotes the Whisper transcript.
  Numbers in parentheses use human transcripts instead.
  The test sets have 843, 266, and 322 samples.}
  \label{tab:existing}
  \vspace{-4mm}
\end{table}
In \cref{tab:gain}, we add the gaze prior and conversational cues to each VLM, separately and together.
The conversational cues alone raise every VLM by 5.0 to 6.7 points, and the gaze prior alone by 7.1 to 10.3 points.
Both together give the largest gain for every VLM, 10.5 to 14.3 points.
Gaze geometry and conversational cues thus play complementary roles, and the gain does not come from the VLM alone.
Qwen3-VL-8B performs best and is our backbone; results with per-cue and zero-shot results are in the supplementary material.

\subsection{Results on Existing Datasets}
\label{sec:existing}

\Cref{tab:existing} tests whether the gains hold with automatic transcripts and on existing datasets.
Replacing human transcripts with Whisper transcripts costs only 0.3 points on SORBET.
On VGS and AVA-LAEO, our model trained on SORBET is applied without further training, and its accuracy rises with the speaker cue and again with the Whisper transcript.
The gaze labels come from the original datasets, and the speaker need not be visible, so the gains do not depend on our annotation or sample selection.
The cues help even though the attended person is the current speaker in only 10.5\% of samples on VGS and 33.2\% on AVA-LAEO, against 60.9\% on the SORBET test set.
On AVA-LAEO, the cross-dataset model reaches 91.0\%, comparable to the in-domain model at 89.4\%.
Conversational cues thus help beyond SORBET, even with automatic transcripts.

\section{Discussion}
\label{discussion}
\paragraph{Limitations and Future Work}
\dataset is built from scripted TV series, where acting and camera framing follow a script.
We use a single keyframe and assume that the observer looks at one of the candidates, so gaze shifts and gaze at objects or off-screen people are not covered.
Mapping names in the transcript to faces requires the character identities of the people in the scene.
On VGS, the base input falls below zero-shot Gaze-LLE, 48.9\% against 51.5\%, so the VLM trained on \dataset does not always transfer.
Raw audio given directly to Qwen2.5-Omni stays below the explicit cues in the supplementary material, and using audio and multiple frames end to end is left for future work.

\section{Conclusion}
\label{sec:conclusion}
We formulated attended person identification in multi-party conversations, where gaze geometry alone becomes ambiguous as more people appear.
We built \dataset with synchronized speech, transcripts, and attended person labels, and proposed \method, which combines gaze geometry with conversational cues in a single VLM.
\method outperforms gaze following, social gaze, and VLM baselines, and conversational cues help across VLMs, with automatic transcripts, and on existing datasets.
Additional visual cues, in contrast, do not help.
The \dataset annotations and our code will be made publicly available.
We hope \dataset and \method lead to gaze models that listen as well as look.

\FloatBarrier

{
    \small
    \bibliographystyle{ieeenat_fullname}
    \bibliography{main}
}

\end{document}